\documentclass[a4paper]{article}

\newcommand{\keywords}[1]{\par\vspace{0.5em}\noindent\textbf{Keywords:} #1}
\usepackage[T1]{fontenc}
\usepackage{graphicx}
\usepackage{mathtools}%
\usepackage{amsmath,amssymb,amsfonts}%
\usepackage{amsthm}
\DeclareMathOperator{\Tr}{Tr}%
\usepackage{enumerate}
\usepackage{crossreftools}
\usepackage{algorithm}
\usepackage{bm}
\usepackage{dsfont}
\usepackage{fancyhdr}
\usepackage{mathrsfs} 
\usepackage{xcolor} 
\usepackage{bbding}  
\usepackage[hidelinks]{hyperref}%
\usepackage{cleveref}%
\newtheorem{assumption}{Assumption}
\newtheorem{prop}{Proposition}%
\newtheorem{definition}{Definition}
\newtheorem{lemma}{Lemma}
\newtheorem{theorem}{Theorem}
\newtheorem{remark}{Remark}
\newtheorem{corollary}{Corollary}
\newcommand{\mygraphic}[1]{\includegraphics[height=#1]{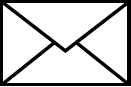}}
\newcommand{\myenv}{(\raisebox{0pt}{\mygraphic{.6em}})}

\newcommand{\orcidl}{\raisebox{-0.0ex}{\href{https://orcid.org/0000-0002-4811-6585}{\includegraphics[height=8pt]{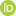}}}}
\newcommand{\orcidlo}{\raisebox{-0.0ex}{\href{https://orcid.org/0000-0003-3096-2844}{\includegraphics[height=8pt]{orcid_logo.png}}}}
\newcommand{\orcidlog}{\raisebox{-0.0ex}{\href{https://orcid.org/0000-0002-0650-3157}{\includegraphics[height=8pt]{orcid_logo.png}}}}
\newcommand{\orcidlogo}{\raisebox{-0.0ex}{\href{https://orcid.org/0000-0001-9623-8053}{\includegraphics[height=8pt]{orcid_logo.png}}}}
\newcommand{\reDEm}{\mb{E}_{\mm{Rad}^{mn},\nu_X^n}} %
\newcommand{\RE}{\mb{E}_{\mm{Rad}^{mn}}} %
\newcommand{\Rad}{\mf{R}_n^m} 
\newcommand{\eR}{\widehat{\mf{R}}_n^m} 
\newcommand{\bsigma}{\bm{\sigma}} 
\newcommand{\bomega}{\bm{\omega}} 
\definecolor{myblue}{RGB}{66, 139, 202}
\newcommand{\E}{\mb{E}} 
\newcommand{\mbS}{\mb{S}_+^m}
\newcommand{\mBO}{\m{B}} 
\newcommand{\mK}{\m{K}}
\newcommand{\N}{\mb{N}} 
\newcommand{\R}{\mb{R}} 
\newcommand{\Rnn}{\R^{\ge 0}} 
\renewcommand{\d}{\mm{d}} 
\renewcommand{\b}{\mathbf} 
\newcommand{\T}{\top} 

\newcommand{\mD}{\m{D}} 
\newcommand{\mY}{\m{Y}} 
\newcommand{\B}{\mb{B}} 
\newcommand{\mR}{\m{R}} 
\newcommand{\C}{\mb{C}} 
\newcommand{\F}{\m{F}} 
\newcommand{\mP}{\m{P}} 
\newcommand{\mX}{\m{X}} 
\newcommand{\bH}{\m{B}(\R^m)} 
\newcommand{\mH}{\m{H}} 
\newcommand{\mB}{\m{B}} 
\newcommand{\GK}{\boldsymbol{\mm{K}}} 
\newcommand{\Gk}{\boldsymbol{\mm{k}}} 
\newcommand{\ra}{\operatorname{ran}} 
\renewcommand{\O}{\mathcal{O}} 

\newcommand{\mf}[1]{\mathfrak{#1}}
\newcommand{\mb}[1]{\mathbb{#1}}
\newcommand{\m}[1]{\mathcal{#1}}

\newcommand{\mm}[1]{\mathrm{#1}}

\allowdisplaybreaks

\begin{document}

\title{Operator-Based Generalization Bound for Deep
Learning: Insights on Multi-Task Learning}
%
%
\author{Mahdi Mohammadigohari \inst{1}\textsuperscript{\myenv}\orcidl \and
Giuseppe Di Fatta\inst{1}\orcidlo\and
Giuseppe Nicosia\inst{3}\orcidlog\and
Panos M Pardalos\inst{4}\orcidlogo}

%
\author{
Mahdi Mohammadigohari\,\orcidl \quad
Giuseppe Di Fatta\,\orcidlo\\[2pt]
\textit{Free University of Bozen-Bolzano, Italy}\\
\href{mailto:mahdi.mohammadigohari@student.unibz.it}{mahdi.mohammadigohari@student.unibz.it} \quad
\href{mailto:giuseppe.difatta@unibz.it}{giuseppe.difatta@unibz.it}
\and
Giuseppe Nicosia\,\orcidlog\\[2pt]
\textit{University of Catania, Italy}\\
\href{mailto:giuseppe.nicosia@unict.it}{giuseppe.nicosia@unict.it}
\and
Panos M. Pardalos\,\orcidlogo\\[2pt]
\textit{University of Florida, USA}\\
\href{mailto:pardalos@ise.ufl.edu}{pardalos@ise.ufl.edu}
}

\date{}   
\maketitle 
%
\thispagestyle{fancy}
\fancyhf{}
\chead{%
  \footnotesize\emph{
  Accepted for LOD 2025. To appear in Lecture Notes in Computer Science (LNCS).
  }
}

\begin{abstract}
This paper presents novel generalization bounds for vector-valued neural networks and deep kernel methods, focusing on multi-task learning through an operator-theoretic framework. Our key development lies in strategically combining a Koopman-based approach with existing techniques, achieving tighter generalization guarantees compared to traditional norm-based bounds. To mitigate computational challenges associated with Koopman-based methods, we introduce sketching techniques applicable to vector-valued neural networks. These techniques yield excess risk bounds under generic Lipschitz losses, providing performance guarantees for applications including robust and multiple quantile regression. Furthermore, we propose a novel deep learning framework—deep vector-valued reproducing kernel Hilbert spaces (vvRKHS)—leveraging Perron-Frobenius (PF) operators to enhance deep kernel methods. We derive a new Rademacher generalization bound for this framework, explicitly addressing underfitting and overfitting through kernel refinement strategies. This work offers novel insights into the generalization properties of multi-task learning with deep learning architectures, an area that has been relatively unexplored until recent developments. 
\keywords{Generalization bounds \and Deep learning \and Kernel methods \and Koopman-based methods \and Multi-task learning \and Perron-Frobenius operators \and Rademacher complexity \and Vector-valued reproducing kernel Hilbert space (vvRKHS).}
\end{abstract}
\section{Introduction}\label{sec:intro}
Despite the remarkable empirical successes of deep learning across diverse fields \cite{Crawshaw2020survey,di_fatta_multi_2023}, a significant gap remains in our theoretical understanding of their generalization capabilities, especially in multi-task learning scenarios. This work is specifically motivated by the need to overcome the limitations of existing generalization bounds in capturing the performance of deep learning models and deep kernel methods in multi-task settings, an area where a deeper understanding of network structure, operator theory, and adaptive kernel refinement strategies is essential for realizing robust and reliable performance. Traditionally distinct, kernel methods and deep neural networks are now being actively investigated for their interconnections. A key perspective is deep kernel learning \cite{ober2021promises}, where composite functions in reproducing kernel Hilbert spaces (RKHSs) are learned from data, and representer theorems ensure data-dependent solutions \cite{bohn2019representer,laforgue2019autoencoding}. This blends deep neural network flexibility with the theoretical rigor of kernel methods. Relatedly, the neural tangent kernel \cite{chen2021deep,jacot2018neural} and convolutional kernel networks \cite{mairal2014convolutional} enable kernel-based analysis of neural networks, and kernel-inspired regularization has been explored for deep learning \cite{bietti2019kernel}.

 The generalization properties of both kernel methods and deep neural networks have been extensively investigated, with Rademacher complexity \cite{mohri2018foundations} serving as a central tool for bounding generalization errors. For kernel methods, Rademacher complexity bounds can be derived directly through the reproducing property. Extensions of this approach have produced generalization bounds for deep kernel methods and vector-valued reproducing kernel Hilbert spaces (vvRKHSs) \cite{huusari2021entangled,laforgue2019autoencoding,sindhwani2013scalable}. Complementing this, a significant body of research has focused on generalization bounds specific to deep neural networks \cite{bartlett2017spectrallynormalized,golowich2018size,hashimoto2023deep,hashimoto2024koopmanbased,ju2022robust,neyshabur2015normbased,suzuki2020compressionbased}. A recent line of research \cite{bartlett2020benign,mallinar2022benign} has focused on analyzing generalization performance through the lens of benign overfitting. Despite these advances, a dedicated generalization analysis method for deep vector-valued kernel learning and multi-output deep neural networks remains underdeveloped, hindering the derivation of tighter, more meaningful generalization error bounds. In the general vector-valued setting, these spaces exhibit several intriguing properties that shed light on multi-task learning with neural networks; such insight cannot be achieved by the former studies. 

This paper introduces novel generalization bounds for vector-valued NNs in the multi-task learning setting. In \cite{Mohammadigohari2025koopman}, we initiated an operator-theoretic approach for analyzing generalization in multi-task deep learning, focusing on full-rank weights and leveraging Koopman operators. We derived a tighter bound compared to the existing Koopman-based bound \cite{hashimoto2024koopmanbased} by applying a new function space. In this work, the first part continues this line by introducing input space sketching—a dimensionality reduction technique to enhance practical applicability in large-scale settings. Drawing inspiration from Kernel Autoencoders (KAEs) and leveraging the framework of reproducing kernel Hilbert $C^*$-modules---a $C^*$-algebra-based extension of RKHS theory---\cite{hashimoto2023deep} pioneered deep RKHM, a deep architecture built by composing functions within RKHMs using the Perron-Frobenius (PF) operator. We strategically combine this approach with existing techniques and, in the second part, we presents a new network architecture, deep vvRKHS, constructed through using PF operators. This architecture loosens assumptions on weight matrices and activation functions from previous models in \cite{Mohammadigohari2025koopman,hashimoto2024koopmanbased} by refining the operator product representation of network layers, resulting in a more compact and expressive structure.

By taking these points into consideration, to enhance generalization performance via our approach, we make the following contributions:
\begin{itemize}
    \item[\textbullet] \textbf{Layer-wise Generalization Analysis.} Our framework enables a layer-wise analysis of generalization by combining our new bound for the initial layers with existing bounds for the subsequent layers, resulting in a tighter overall estimate that accounts for the specific role of each layer (see \Cref{Theo:koopmanbdintegration} \& \Cref{rem:combinedKoopman}).

     \item[\textbullet] \textbf{Deep Sketched Input Kernel Regression.} We introduce Deep Sketched Input Kernel Regression, generalizing input sketching to vector-valued neural networks (see \Cref{subsec:p_sparsified_estimator}).
        \item[\textbullet] \textbf{Deep vvRKHS Framework.} This work introduces deep vvRKHS, a framework that enhances deep kernel methods. By using PF operators, we derive new generalization bounds and provide a connection to underfitting or overfitting phenomena (see \Cref{rem: deepvvRKHS_with_sep_ker_refine}).
\end{itemize}
The paper is structured as follows: \Cref{sec:MathPrelimi} reviews necessary mathematical background and notation. \Cref{sec:KoopmanBasedDNN} presents our operator representation framework for deep learning using Koopman operators. \Cref{sec:rad_compl_bounds_DNNs} establishes generalization bounds on Rademacher complexity and excess risk bounds for sketched  estimator. \Cref{sec:operator_vvRKHS} then introduces our deep vvRKHS framework, a new generalization bound, and connections to underfitting or overfitting. Finally, \Cref{sec6} concludes the paper with a discussion of results and potential future directions.
\subsection{Related works}\label{subsec:Reworks}
\textbf{Multi-Task Learning:} The advantages of multi-task learning have been extensively explored in the machine learning literature \cite{argyriou2006multitask}. Recent theoretical research has focused on properties of multi-task neural networks and how shared representations and task relationships influence generalization \cite{lindsey2023implicit,collins2024provable,shenouda2024variation}. Initial generalization results in multi-task learning include Rademacher complexity-based bounds for linear classification \cite{Maurer2006Bounds}, subsequently refined to derive risk estimates in trace norm regularized models \cite{Pontil2013Excess}. More recently, sharper risk bounds leveraging local Rademacher complexity analysed the role of common regularization strategies \cite{Yousefi2018Local}. To address the current scarcity of multi-task deep learning theories, this work introduces a novel framework to analyzing the generalization characteristics of functions learned by vector-valued deep learning architectures with use of transfer operators, providing new insights into multi-task learning with neural networks.\\[4pt]
\noindent \textbf{Sketching:} To alleviate the computational demands of kernel methods, sketching techniques have emerged as a powerful tool for approximating kernel matrices and reducing memory requirements \cite{mahoney2011randomized,woodruff2014sketching,yang2017randomized}. By employing randomized linear projections, sketching can enable near-optimal nonparametric regression \cite{yang2017randomized} and provide efficient solutions for large-scale problems. Approaches such as Random Fourier Features \cite{li2021towards} have proven effective, while recent work on $p$-sparsified sketches \cite{elahmad2023fast} demonstrates that structured sparsity can offer both computational advantages and strong theoretical guarantees, extending the range of applicability for generic Lipschitz losses. The core aim is the same as well known by \cite{caponnetto2007optimal}.

\section{Mathematical Preliminaries}\label{sec:MathPrelimi}
This section presents our notation (\Cref{sec:notations}) and the required mathematical background (\Cref{sec:background}).
\subsection{Key notations and definitions}\label{sec:notations}
To ensure clarity and consistency throughout this work, we establish the following notation. We denote the set of non-negative real numbers by $\Rnn$. For a function $p$ belonging to $L^{\infty}\left(\R^d\right)$, its $L^{\infty}$-norm is written as $\|p\|_{L^\infty}$. For $p \ge 1$ and any $f = (f_1, \dots, f_m) \in L^2(\R^d; \R^m)$, the $L^p$-norm over $\R^d$ satisfies $\|f\|_{L^p(\R^d; \R^m)}^2 = \sum_{j=1}^m \int_{\R^d} \|f_j(\b x)\|^p \d \b x = \sum_{j=1}^m \|f_j\|_{L^p\left(\R^d\right)}^p$, where $f_j$ is the $j$-th component of $f$, and $\|f_j\|_{L^p\left(\R^d\right)}$ is the standard Lebesgue $L^p$-norm of $f_j$ over $\R^d$. In the case of $p = 2$ the Lebesgue space is a Hilbert space. Given a function $f \in L^2(\R^d,\C^m)$, we define its Fourier transform as $\hat{f}(\bomega) = \frac{1}{(2\pi)^{d/2}} \int_{\R^d} f(\b x) e^{i \b x^\T \cdot \bomega} \d \b x$.

A Hilbert space is a mathematical structure consisting of a vector space endowed with an inner product, denoted by $\langle \cdot, \cdot \rangle$, which in turn induces a norm specified by $\| \cdot \| \coloneqq \sqrt{\langle \cdot, \cdot \rangle}$. Let $\mX$ be a Hilbert space. The notation $\mBO(\mX)$ represents the space of bounded linear operators on $\mX$, while $\mBO(\mX)^+$ denotes the subset of positive bounded linear operators. For a linear operator $\b W$ acting on a Hilbert space $\mX$, we denote its range and kernel by $\ra(\b W)$ and $\ker(\b W)$, respectively.

A kernel is a symmetric function $k : \mX \times \mX \to \R$ such that, for any $n \in \N$, any $(\b x_i, \alpha_i)_{i=1}^n \in (\mX \times \R)^n$, $\sum_{i,j=1}^n \alpha_i k(\b x_i, \b x_j) \alpha_j \geq 0$ holds. There exists a one-to-one correspondence between kernels and reproducing kernel Hilbert spaces (RKHS). The RKHS $\mH_k \subset \R^{\mX}$ is a Hilbert space of functions such that, for every $\b x \in \mX$, the function $k_\b x \coloneqq k(\cdot, \b x)$ belongs to $\mH_k$. Moreover, for all $h \in \mH_k$ and $\b x \in \mX$, $h(\b x) = \langle h, k_\b x \rangle_{\mH_k}$\footnote{Here, $k(\cdot, \b x): \mX \to \R$ is defined by $\b x' \mapsto k(\b x',\b x)$ with $x$ fixed.}.

Let $\mY$ be a Hilbert space. An operator-valued kernel (OVK) is a function $K : \mX \times \mX \to \mBO(\mY)$, such that for any $n \in \mathbb{N}$, any $(\b x_i, \b y_i)_{i=1}^n \in (\mX \times \mY)^n$, $\sum_{i,j=1}^n \b y_i^\T K(\b x_i, \b x_j) \b y_j \geq 0$ and $K(\b x,\b x') = K(\b x',\b x)^\T$ holds for all $(\b x,\b x')\in \mX^2$. There exists a one-to-one correspondence between MVKs and vvRKHSs; the vvRKHS $\mH_K \subset (\R^m)^{\mX}$ is a Hilbert space of functions such that, for every $\b x \in \mX$ and $\b y \in \R^m$, the function $K_\b x y \coloneqq K(\cdot, \b x)y$ belongs to $\mH_K$, and for all $h \in \mH_K$, $\b x \in \mX$, and $\b y \in \R^d$, $h(\b x)^\T \b y = \langle h, K_\b x \b y \rangle_{\mH_K}$. For any $\b x \in \mX$, $\phi(\b x) \coloneqq K(\cdot,\b x)$ denotes the feature map.

Given vvRKHSs $\m{H}_{K_1}$ and $\m{H}_{K_2}$ on $\R^{d_1}$ and $\R^{d_2}$ with MVKs $K_1$ and $K_2$, the Koopman operator $\mK_f : \m{H}_{K_2} \to \m{H}_{K_1}$ for $f : \R^{d_1} \to \R^{d_2}$ is defined as $\mK_f g = g \circ f$ for $g \in \m{D}_f$, where $\m{D}_f= \{g \in \m{H}_{K_2} \mid g \circ f \in \m{H}_{K_1}\}$ and $\circ$ is the composition operator. 

Given $\mY$-valued RKHSs $\m{H}_{K_1}$ and $\m{H}_{K_2}$ on $\mX_1$ and $\mX_2$, respectively. For a mapping $f$ from $\mX_1$ to $\mX_2$, the PF operator associated with $f$, denoted $\mP_f$, is a linear transformation from $\mH_{K_1}$ to $\mH_{K_2}$ defined by the property that $\mP_f \phi_1(\b x) \b y = \phi_2(f(\b x)) \b y$ for all $\b x \in \mX_1$ and $\b y \in \mY$.

Given two $\mBO(\mY)$-valued reproducing kernels, $K_1$ and $K_2$, defined on $\mX$, $\mH_{K_1} \leq \mH_{K_2}$ implies that $\mH_{K_1} \leq \mH_{K_2}$, and for all $f \in \mH_K$, $\|f\|_{\mH_{K_1}} = \|f\|_{\mH_{K_2}}$. $K_2$ is termed a refinement of $K_1$ when $\mH_{K_1} \subseteq \mH_{K_2}$ holds. This refinement is only considered nontrivial if $K_1$ and $K_2$ are not the same. For $\b A, \b B \in \mBO(\R^m)$, we write $\b A \leq \b B$ if $\b B - \b A$  positive semi-definite.

The unit ball in a Hilbert space $\mH$, denoted by $\B_1(\mH)$, is defined as the set of all elements $f \in \mH$ with norm $\|f\|_{\mH} \leq 1$. The restriction of a function $h$ defined on $\R^d$ to a subset $S \subseteq \R^d$ is denoted by $h|_S$. Finally, the transpose of a matrix $\b W$ is represented by $\b W^\T$. With $\mbS \subset \R^{m \times m}$ we denote  the set of $m \times m$ symmetric and positive semi-definite (p.s.d.) matrices. For a matrix $\b A \in \R^{m \times p}$, let $\b A_{i:}$ denotes its $i$-th row in $\R^p$, and $\b A_{:j}$ its $j$-th column in $\R^m$. The $d \times d$ identity matrix is represented by $\b I_d$. Given $\mH \subset \mY^\mX$ as a hypothesis set and random variables $(X, Y) \in \mX \times \mY$ with distribution $P$. For a function $f : \mX \to \mY$, we define expected risk (true risk), $\mR(f) = \E_{(X, Y) \sim P}[\ell(f(X), Y)]$, and $\mR_n(f) = \frac{1}{n} \sum_{i=1}^n \ell(f(\b x_i), \b y_i)$ for any function $\ell : \mY \times \mY \to \R$. 

\subsection{Vector-valued Sobolev spaces}\label{sec:background}
In this work, we employ vvRKHSs as our function spaces. Given $d, m \in \N$, $s \ge 0$, and $s > d/2$, we define the Hilbert space 
\begin{align*}
H^s(\R^d,\R^m) \coloneqq \left\{f \in L_2(\R^d,\R^m) \mid (1 + \|\bomega\|_2^2)^s \hat{f} \in L_2(\R^d,\R^m)\right\}.    
\end{align*}
This space is equipped with the inner product 
\begin{align*}
\langle f,g \rangle_{H^s(\R^d,\R^m)} = \int_{\R^d} (1 + \|\bomega\|_2^2)^s \hat{f}(\bomega)^*\hat{g}(\bomega)\mm{d}\bomega,
\end{align*}
where symbols $\hat{\cdot}$, $*$ denote the Fourier transform and the complex conjugate (or Hermitian transpose), respectively. The reproducing kernel for $H^s(\R^d,\R^m)$ is $K_s(\b x, \b y) \coloneqq k_s \b M$, where $k_s$ can be chosen as a radial ($\textit{i.e.}$, it can be written in the form $k_s(\b x,\b y) = \phi_s(\|\b x-\b y\|_2)$ with a function $\phi_s : [0, \infty ) \to \R$.) and $\b M \in \mbS$. A consequence of $H^s(\R^d,\R^m)$ being a subset of $L^2(\R^d,\R^m)$ is that all functions in $H^s(\R^d,\R^m)$ tend to zero at infinity. For $s > d/2$, $H^s(\R^d,\R^m)$ is the Sobolev space $W^{s,2}(\R^d,\R^m)$. Details regarding Sobolev spaces of vector-valued functions and their associated vvRKHS can be found in \cite{wittwar2022,li2024}.

\begin{figure}[htbp]
  \centering
  \includegraphics[width=0.8\linewidth]{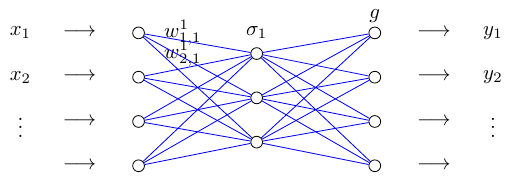}
  \caption{Illustration of the proposed network architecture (adapted from \cite{Mohammadigohari2025koopman}, Figure 1). The network consists of an input layer, one hidden layer, activation function $\sigma_1$, final nonlinear transformation $g$, and an output layer.}
  \label{fig:myfig1}
\end{figure}

\subsection{Vector-valued Rademacher complexity}
We study the general setting of multi-output (vector-valued) supervised learning.  
Consider a training dataset $\mD_{XY,n} = \{(\b x_i, y_i)\}_{i=1}^n \sim \nu_{XY}^n$, where  
(i) $\nu_{XY} \in \mP(\mX \times \mY)$ denotes the joint distribution describing the dependence between inputs $X$ and outputs $Y$;  
(ii) $\mX \subset \R^d$ and $\mY \subset \R^m$ denote the input and output spaces, respectively; and  
(iii) $\nu_X$ is the marginal distribution of $X$.\footnote{The marginal $\nu_X$ will be used in the definition of the expected Rademacher complexity.}  
The goal of learning is to construct a predictor $f:\mX \to \mY$ from the sample $\mD_{XY,n}$ such that $f(x)$ accurately predicts the output $y$ for unseen data points $x$.

To quantify the richness of a class of vector-valued functions, we extend the classical notion of Rademacher complexity to the multi-output setting as follows.

\begin{definition}[(empirical) vector-valued Rademacher complexity]
Let $\F$ be a hypothesis class consisting of functions $f:\mX \to \R^m$ defined on $\mX$.  
Let 
\begin{align*}
\bsigma_i = (\sigma_{i1},\ldots,\sigma_{im}) \sim \mathrm{Rad}^m, 
\quad i \in [n],
\end{align*}
be independent Rademacher vectors, where each component $\sigma_{ij}$ is an independent random variable uniformly distributed on $\{-1,+1\}$.  
For a fixed dataset $\mD_n = \{\b x_i\}_{i=1}^n \subset \mX$, the empirical vector-valued Rademacher complexity of $\F$ is defined as
\begin{align}
\eR(\F)
\coloneqq
\RE\!\left[
\sup_{f\in\F}
\frac{1}{n}\left|\sum_{i=1}^n
\langle \bsigma_i, f(\b x_i)\rangle\right|
\right],
\label{eq:empirical-Rad-vector}
\end{align}
where $\langle\cdot,\cdot\rangle$ denotes the Euclidean inner product in $\R^m$.  
Taking the expectation with respect to $(X_i)_{i=1}^n \sim \nu_X^n$ yields the (expected) vector-valued Rademacher complexity:
\begin{align}
\Rad(\F)
\coloneqq
\RE\!\left[\eR(\F)\right]
=
\reDEm\!\left[
\sup_{f\in\F}
\frac{1}{n}\left|\sum_{i=1}^n
\langle \bsigma_i, f(X_i)\rangle\right|
\right].
\label{eq:expected-Rad-vector}
\end{align}
Equivalently, we may write $\bsigma = (\bsigma_1,\ldots,\bsigma_n)\sim \mathrm{Rad}^{nm}$.
\end{definition}

\section{Operator Representation framework for deep learning}\label{sec:KoopmanBasedDNN}
This work analyzes the generalization capabilities of $L$-layer deep neural networks mapping from an input space $\R^{d_0}$ to an output space $\R^m$. The network architecture is formally defined as:
\begin{align}\label{PS:mainnetfun}
f = g \circ \b b_L \circ \b W_L \circ \sigma_{L-1} \circ \b b_{L-1} \circ \b W_{L-1} \circ \dots \circ \sigma_1 \circ \b b_1 \circ \b W_1 
\end{align}
comprising weight matrices $\b W_l \in \R^{d_l \times d_{l-1}}$ ($l = 1, \dots, L$) that map from $\R^{d_{l-1}}$ to $\R^{d_l}$ (assumed injective or invertible), shift operators $\b b_l$ ($l = 1, \dots, L$) defined as $\b b_l(\b x) = \b x + \b a_l$ with bias vectors $\b a_l \in \R^{d_l}$, nonlinear activation functions $\sigma_l : \R^{d_l} \rightarrow \R^{d_l}$ ($l = 1, \dots, L - 1$), and a final nonlinear transformation $g : \R^{d_L} \rightarrow \R^m$ mapping the output to the $m$-dimensional space.

Building upon the operator-theoretic framework developed in \cite{Mohammadigohari2025koopman,hashimoto2024koopmanbased}, we present an in-depth analysis of vector-valued neural networks through the lens of vvRKHSs. With $d_0$ representing the input dimension and $d_l$ the width of layer $l$ ($l = 1, \dots, L$), the $L$-layer network $f$ can be expressed using Koopman operators as
\begin{align}\label{ps:koopmanform}
f = \mK_{\b W_1} \mK_{\b b_1} \mK_{\sigma_1} \dots \mK_{\b W_{L-1}} \mK_{\b b_{L-1}} \mK_{\sigma_{L-1}} \mK_{\b W_L} \mK_{\b b_L} g, 
\end{align}
where $\mK_{\b W_l} : H^{s_l}\left(\R^{d_l}, \R^m\right) \rightarrow H^{s_{l-1}}\left(\R^{d_{l-1}}, \R^m\right)$, and $\mK_{\b b_l}, \mK_{\sigma_l} : H^{s_l}\left(\R^{d_l}, \R^m\right) \rightarrow H^{s_l}\left(\R^{d_l}, \R^m\right)$. Details of the shallow network's architecture are depicted in Figure \ref{fig:myfig1}.

In this paper, we restrict our analysis to the finite-dimensional setting, where the output space $\mY$ is a subset of $\R^m$ with $m \ge 2$, and the matrix-valued kernel satisfies $K(\b x,\b x') \in \R^{m \times m}$ for all $\b x,\b x' \in \mX$. Given a training sample ${\b x_1, \ldots, \b x_n}$, the Gram matrix is defined as $\GK = (K(\b x_i, \b x_j))_{1 \leq i,j \leq n} \in \R^{nm \times nm}$. We further assume a decomposable kernel structure, such that $K(\b x,\b x') = k(\b x,\b x')\b M$ for some scalar kernel $k$ and positive semidefinite matrix $\b M \in \R^{m \times m}$. Consequently, the Gram matrix can be expressed as $\GK = \Gk \otimes \b M$, where $\Gk \in \R^{n \times n}$ is the scalar Gram matrix and $\otimes$ denotes the Kronecker product.

We make the following assumptions we will use throughout this paper.
\begin{assumption} \label{assum:true_risk_min}
The true risk is minimized at $f_{H^{s_0}} \coloneqq f_{H^{s_0}\left(\R^{d_0}, \R^m\right)}$.
\end{assumption}
\begin{assumption} \label{assum:hypothesis_set}
The hypothesis set considered is $\B_1\left(H^{s_0}\left(\R^{d_0}, \R^m\right)\right)$.
\end{assumption}
\begin{assumption} \label{assum:lipschitz}
For all $y \in \mY$, the mapping $z \mapsto \ell(z, y)$ is $J_{\ell}$-Lipschitz.
\end{assumption}
\begin{assumption} \label{assum:kernel_bound}
For all $\b x,\b x' \in \mX$, we have $k_{s_0}(\b x,\b x') \leq \kappa$.
\end{assumption}
\begin{assumption}\label{assum:final_transformation}
The final nonlinear transformation $g$ is contained in $H^{s_L}(\R^{d_{L}},\\ \R^m)$, and $\mK_{\sigma_l}$ are bounded for $l = 1,\ldots, L-1$.   
\end{assumption}
We denote by $\F$ the set of all functions in the form of \eqref{PS:mainnetfun} with \Cref{assum:final_transformation}. Let $g(\b x) = e^{-\|\b x\|_2^2}$ for $\b x \in \R^{d_L}$ and $s_L > d_L/2$. Then, $g(\b x) = \b M e^{-\|\b x\|_2^2} c^{\T}$ satisfies this assumption for $\b M \in \mbS$ and any $\b c \in \R^m$.

Following standard practice in kernel methods and deep learning theory, our analysis relies on a set of key assumptions. Specifically, Assumption~\ref{assum:true_risk_min}, the attainability condition, is a common starting point in kernel literature \cite{caponnetto2007optimal,rudi2017generalization,li2021towards}, positing the existence of a risk minimizer within our hypothesis space. Assumption~\ref{assum:hypothesis_set} addresses complexity control by restricting the hypothesis space to a unit ball, allowing us to normalize without loss of generality. Assumption~\ref{assum:lipschitz}, the Lipschitz continuity of the loss function, is a classical requirement for bounding generalization error. Finally, Assumption~\ref{assum:final_transformation}, pertaining to the properties of the final nonlinear transformation, is justified by Remark 1 and Lemma 1 of \cite{Mohammadigohari2025koopman}. These assumptions, while potentially restrictive, enable us to establish meaningful theoretical guarantees.

\section{Generalization bounds via Rademacher complexity}\label{sec:rad_compl_bounds_DNNs}
\label{sec4}
We proceed by bounding the Rademacher complexity, analyzing the case where weight matrices are invertible or injective.

\subsection{Upper bound for injective weight matrices}\label{sub:2}
To extend the analysis, we now consider the setting of injective weight matrices. Given constants $C, D > 0$, we define the set of weight matrices $\m{W}_l^{\,\scriptscriptstyle C,D}$ as
 \begin{align*}
\m{W}_l^{\,\scriptscriptstyle C,D} = \left\{\b W\in \R^{d_{l-1}\times d_l} \mid d_l\ge d_{l-1} \,, \|\b W\| \leq C,\, \det\left(\b W^{\T}\b W\right)^{1/2} \ge D\right\},     
 \end{align*}

\noindent where the condition $d_l \ge d_{l-1}$ ensures injectivity. Let $\F_{\scriptscriptstyle \mm{inj}}$ represent the set of the functions whose weight matrices $\b W_l$ belong to $\m{W}_l^{\,\scriptscriptstyle C,D}$. Defining $f_l = g  \circ \, \b b_L  \circ \, \b W_L  \circ \, \sigma_{L-1} \circ \, \b b_{L-1}  \circ \, \b W_{L-1}  \circ \, \ldots  \circ \, \sigma_l  \circ \, \b b_l$ and the ratio of norms $\mm{G}_l =\frac{\|f_l|_{\ra(\b W_l)}\|_{H^{s_{l-1}}\left(\ra(\b W_l), \R^m\right)}}{\|f_l\|_{H^{s_l}\left(\R^{d_l},\R^m\right)}}$, the following theorem provides a bound on the Rademacher complexity for $\F_{\scriptscriptstyle \mm{inj}}$. Using a similar argument as in the proof of Theorem $3$, and applying inequality $(10)$ from Theorem $2$ in \cite{Mohammadigohari2025koopman}, yields the following results.
\begin{lemma}
\label{Corolla:upperbound_inj_weight}
The Rademacher complexity $\eR(\F_{\scriptscriptstyle \mm{inj}})$ is bounded as
 \begin{align}\label{eq77}
 \eR(\F_{\scriptscriptstyle \mm{inj}}) \leq &\|g\|_{H^{s_L}\left(\R^{d_L},\R^m\right)}\sqrt{\frac{\kappa}{n}\Tr(\b M)}\sup_{\b W_l \in \m{W}_l^{\,\scriptscriptstyle C,D}} \prod_{l = 1}^L \mm{G}_l \nonumber\\
 &\cdot \sup_{\bomega \in \ra(\b W_l)} \left|\frac{1 + \|\b W_l^{\T} \bomega\|^2_2}{1 + \|\bomega\|^2_2}\right|^{s_{l-1}/2} \frac{1}{\det(\b W_l^{\T}\b W_l)^{1/4}}\prod_{l = 1}^{L-1}\|\mK_{\sigma_l}\|.
\end{align}
\end{lemma}

\subsection{Integrating the Koopman-based bound}\label{subsec:combining_koopman}
The Koopman-based bound is flexible and can be combined with other bounds, as demonstrated previously (\cite{hashimoto2024koopmanbased}, Section 4.4 and Proposition 19). This approach can be extended to our setting using the same technique. Specifically, the complexity of an $L$-layer, multi-output network can be decomposed into the Koopman-based bound for the first $l$ layers and the bound for the remaining $L - l$ layers. Furthermore, the overall bound includes a term that quantifies the approximation capability of the function class associated with these $L - l$ layers. This provides a new perspective, allowing us to integrate our bound with existing bounds by applying our method to lower layers and existing bounds to higher layers. Consequently, this enables us to derive a tight overall bound that accounts for the contribution of each individual layer. Let $\F^{\,1 \mathord{:} l'}$ be the set of all functions in the form
\begin{align}
\label{MTLRe}
 \sigma_{l'}  \circ \, \b b_{l'}  \circ \, \b W_{l'}  \circ \, \sigma_{l'-1}  \circ \, \b b_{l'-1} \circ \, \b W_{l'-1}  \circ \, \cdots  \circ \, \sigma_1  \circ \, \b b_1  \circ \, \b W_1,  
\end{align}
for $0\leq l' \leq L$, with Assumption \eqref{assum:final_transformation}, and let $\F^{\,1\mathord{:} l'}_{\scriptscriptstyle \mm{inj}} = \{f \in \F^{\,1 \mathord{:} l'} \,\,\, | \,\,\, \b W_l\in \mathcal{W}_l^{\,\scriptscriptstyle C,D}\}$. For $l' \leq L - 1$, consider any nonempty subset $\F^{\,l'+1 \mathord{:} L}$ of all
functions in $H^{s_{l'}}\left(\R^{d_{l'}},\R^m\right)$ which has the form
\begin{align*}
 g \circ \, \b b_L \circ \, \b W_L  \circ \, \sigma_{L-1}  \circ \, \b b_{L-1} \circ \, \b W_{L-1}  \circ \, \cdots  \circ \, \sigma_{l'+1}  \circ \, \b b_{l'+1}  \circ \, \b W_{l'+1},  
\end{align*}
For $l' = L$, we set $\F^{\,l'+1 \mathord{:} L} = \{g\}$. Let $\F^{\,1 \mathord{:} L}_{\scriptscriptstyle \mm{comb}} = \left\{f_1 \circ\, f_2 \,\,\, | \,\,\, f_1 \in \F^{\,l'+1 \mathord{:} L},\, f_2 \in \F^{\, 1\mathord{:} l'}_{\scriptscriptstyle \mm{inj}}\right\}$. 

The theorem below establishes a relationship between the Rademacher complexities of $\F^{\,1 \mathord{:} L}_{\scriptscriptstyle \mm{comb}}$ and $\F^{\,l'+1 \mathord{:} L}$.
\begin{theorem}
\label{Theo:koopmanbdintegration}
The Rademacher complexity $\eR\left(\F^{\,1 \mathord{:} L}_{\scriptscriptstyle \mm{comb}}\right)$ is bounded as
 \begin{align}
&\eR\left(\F^{\,1 \mathord{:} L}_{\scriptscriptstyle \mm{comb}}\right) \leq \sup_{\substack{\b W_l,\\ l=\{1,2,\cdots,l'\}}} \prod_{l = 1}^{l'} \mm{G}_l\sup_{\bomega \in \mathcal{R}(\b W_l)}\left|\frac{1 + \|\b W_l^{\top} \bomega\|^2_2}{1 + \|\bomega\|^2_2}\right|^{s_{l-1}/2} \nonumber\\
&\qquad \frac{\|\mK_{\sigma_l}\|}{\det(\b W_l^{\top}\b W_l)^{1/4}} \left(\eR\left(\F^{\,l'+1 \mathord{:} L}\right) + \sqrt{\frac{\kappa}{n}\Tr(\b M)} \vphantom{\inf_{{h}' \in \F^{\,l'+1 \mathord{:} L}} \RE^{\frac{1}{2}}\left[\sup_{h'' \in \F^{\,l'+1 \mathord{:} L}}\left\|h' - \frac{\gamma_n\|h''\|_{H^{s_{l'}}\left(\R^{d_{l'}},\R^m\right)}}{{\|\Tilde{\upsilon}}_n\|_{H^{s_{l'}}\left(\R^{d_{l'}},\R^m\right)}}\Tilde{\upsilon}_n\right\|^2_{H^{s_{l'}}\left(\R^{d_{l'}},\R^m\right)}\right]}\right.\nonumber\\
&\left.\inf_{{h}' \in \F^{\,l'+1 \mathord{:} L}} \RE^{\frac{1}{2}}\left[\sup_{h'' \in \F^{\,l'+1 \mathord{:} L}}\left\|h' - \frac{\gamma_n\|h''\|_{H^{s_{l'}}\left(\R^{d_{l'}},\R^m\right)}}{{\|\Tilde{\upsilon}}_n\|_{H^{s_{l'}}\left(\R^{d_{l'}},\R^m\right)}}\Tilde{\upsilon}_n\right\|^2_{H^{s_{l'}}\left(\R^{d_{l'}},\R^m\right)}\right]\right).
\end{align}
\begin{proof}
Let $\Tilde{\mathbf{x}} = (\Tilde{\b x}_1,\cdots,\Tilde{\b x}_n) \in (\R^{d_{l'}})^n$. Let $\upsilon_n(\bomega) = \sum_{i=1}^n\boldsymbol{\sigma}_i(\bomega)K_{s_0}(\cdot,\b x_i)$, $\Tilde{\upsilon}_n(\bomega) = \sum_{i=1}^n\boldsymbol{\sigma}_i(\bomega)K_{s_{l'}}(\cdot,\Tilde{\b x}_i)$, $\gamma_n = \|\upsilon_n\|_{H^{s_0}\left(\R^{d_0},\R^m\right)}/\|\Tilde{\upsilon}_n\|_{H^{s_{l'}}\left(\R^{d_{l'}},\R^m\right)}$, and \begin{align*}\eta_l = \mm{G}_l\sup_{\bomega \in \mathcal{R}(\b W_l)}\left|\frac{1 + \|\b W_l^{\top} \bomega\|^2_2}{1 + \|\bomega\|^2_2}\right|^{s_{l-1}/2} \frac{\|\mK_{\sigma_l}\|}{\det(\b W_l^{\top}\b W_l)^{1/4}}\end{align*}. Then, by the reproducing property of $H^{s_0}\left(\R^{d_0},\R^m\right)$, we have
\begin{align}
&\frac{1}{n}\RE \left[\sup_{\substack{{f \in \F^{\,1 \mathord{:} L}_{\scriptscriptstyle \mm{comb}}}}} \sum_{i=1}^n \langle\boldsymbol{\sigma}_i ,f(\b x_i)\rangle_{\R^m} \right] = \frac{1}{n}\RE \left[\sup_{\substack{{f \in \F^{\,1 \mathord{:} L}_{\scriptscriptstyle \mm{comb}}}}} \sum_{i=1}^n  \langle \upsilon_n,f\rangle_{H^{s_0}\left(\R^{d_0},\R^m\right)} \right]\nonumber\\
 &\leq \frac{1}{n}\RE \left[\sup_{\substack{{f \in \F^{\,1 \mathord{:} L}_{\scriptscriptstyle \mm{comb}}}}} \|\upsilon_n\|_{H^{s_0}\left(\R^{d_0},\R^m\right)}\prod_{l = 1}^{l'} \mm{G}_l \sup_{\bomega \in \mathcal{R}(\b W_l)}\left|\frac{1 + \|\b W_l^{\top} \bomega\|^2_2}{1 + \|\bomega\|^2_2}\right|^{s_{l-1}/2} \vphantom{\cdot \frac{\|\mK_{\sigma_l}\|}{\det(\b W_l^{\top}\b W_l)^{1/4}} \|\mK_{\b W_{l'+1}} \mK_{\b b_{l'+1}}\mK_{\sigma_{l'+1}} \cdots \mK_{\b W_L} \mK_{\b b_L}g\|_{H^{s_{l'}}\left(\R^{d_{l'}},\R^m\right)}}\right.\nonumber\\
&\qquad \qquad \left.\cdot \frac{\|\mK_{\sigma_l}\|}{\det(\b W_l^{\top}\b W_l)^{1/4}} \|\mK_{\b W_{l'+1}} \mK_{\b b_{l'+1}}\mK_{\sigma_{l'+1}} \cdots \mK_{\b W_L} \mK_{\b b_L}g\|_{H^{s_{l'}}\left(\R^{d_{l'}},\R^m\right)}\right]\label{th5proofeq2}\\
  &\leq \frac{1}{n}\RE\left[\sup_{\substack{\b W_l,\\ j=\{1,2,\cdots,L\}}}\prod_{l = 1}^{l'} \eta_l \vphantom{.\cdot\sup_{h'' \in \F^{\,l'+1 \mathord{:} L}} \bigg\langle \Tilde{\upsilon}_n, \frac{\|h''\|_{H^{s_{l'}}\left(\R^{d_{l'}},\R^m\right)} \|\upsilon_n\|_{H^{s_0}\left(\R^{d_0},\R^m\right)}}{{\|\Tilde{\upsilon}_n}\|^2_{H^{s_{l'}}\left(\R^{d_{l'}},\R^m\right)}}\Tilde{\upsilon}_n \bigg\rangle_{H^{s_{l'}}\left(\R^{d_{l'}},\R^m\right)}}\right.\nonumber\\
  &\qquad \left.\cdot\sup_{h'' \in \F^{\,l'+1 \mathord{:} L}} \bigg\langle \Tilde{\upsilon}_n, \frac{\|h''\|_{H^{s_{l'}}\left(\R^{d_{l'}},\R^m\right)} \|\upsilon_n\|_{H^{s_0}\left(\R^{d_0},\R^m\right)}}{{\|\Tilde{\upsilon}_n}\|^2_{H^{s_{l'}}\left(\R^{d_{l'}},\R^m\right)}}\Tilde{\upsilon}_n \bigg\rangle_{H^{s_{l'}}\left(\R^{d_{l'}},\R^m\right)}\right]\nonumber\\
&\leq \sup_{\substack{\b W_l,\\ j=\{1,2,\cdots,L\}}} \prod_{l = 1}^{l'} \eta_l \left(\eR\left(\F^{\,l'+1 \mathord{:} L}\right) + \frac{1}{n}\RE\left[\| \Tilde{\upsilon}_n\|_{H^{s_{l'}}\left(\R^{d_{l'}},\R^m\right)}\vphantom{.\cdot \sup_{h'' \in \F^{\,l'+1 \mathord{:} L}}\left\|\frac{\gamma_n\|h''\|_{H^{s_{l'}}\left(\R^{d_{l'}},\R^m\right)}}{{\|\Tilde{\upsilon}_n}\|_{H^{s_{l'}}\left(\R^{d_{l'}},\R^m\right)}}\Tilde{\upsilon}_n - \nu\right\|_{H^{s_{l'}}\left(\R^{d_{l'}},\R^m\right)}}\right.\right. \nonumber\\
&\qquad \qquad \qquad \qquad \left.\left.\cdot \sup_{h'' \in \F^{\,l'+1 \mathord{:} L}}\left\|\frac{\gamma_n\|h''\|_{H^{s_{l'}}\left(\R^{d_{l'}},\R^m\right)}}{{\|\Tilde{\upsilon}_n}\|_{H^{s_{l'}}\left(\R^{d_{l'}},\R^m\right)}}\Tilde{\upsilon}_n - \nu\right\|_{H^{s_{l'}}\left(\R^{d_{l'}},\R^m\right)}\right]\right)\label{th5proofeq3}\\
&\leq \sup_{\substack{\b W_l,\\ j=\{1,2,\cdots,L\}}} \prod_{l = 1}^{l'} \eta_l \left(\eR\left(\F^{\,l'+1 \mathord{:} L}\right) + \frac{1}{n}\RE^{\frac{1}{2}}[\| \Tilde{\upsilon}_n\|^2_{H^{s_{l'}}\left(\R^{d_{l'}},\R^m\right)}] \vphantom{.\cdot \RE^{\frac{1}{2}}\left[\sup_{h'' \in \F^{\,l'+1 \mathord{:} L}}\left\|\frac{\gamma_n\|h''\|_{H^{s_{l'}}\left(\R^{d_{l'}},\R^m\right)}}{{\|\Tilde{\upsilon}_n}\|_{H^{s_{l'}}\left(\R^{d_{l'}},\R^m\right)}}\Tilde{\upsilon}_n -\nu\right\|^2_{H^{s_{l'}}\left(\R^{d_{l'}},\R^m\right)}\right]}\right.\nonumber\\
&\qquad \qquad \qquad \left.\cdot \RE^{\frac{1}{2}}\left[\sup_{h'' \in \F^{\,l'+1 \mathord{:} L}}\left\|\frac{\gamma_n\|h''\|_{H^{s_{l'}}\left(\R^{d_{l'}},\R^m\right)}}{{\|\Tilde{\upsilon}_n}\|_{H^{s_{l'}}\left(\R^{d_{l'}},\R^m\right)}}\Tilde{\upsilon}_n -\nu\right\|^2_{H^{s_{l'}}\left(\R^{d_{l'}},\R^m\right)}\right]\right)\label{th5proofeq4}
\end{align}
where \eqref{th5proofeq2} is derived from the Cauchy–Schwartz inequality and in the same manner as the proof of \Cref{Corolla:upperbound_inj_weight}, \eqref{th5proofeq3} is again a consequence of the Cauchy–Schwartz inequality, \eqref{th5proofeq4} results from the Jensen's inequality. Since $\nu \in \F^{\,l'+1 \mathord{:} L}$ is arbitrary, we reach the desired result. \hfill$\square$
\end{proof}
\end{theorem}
Note that \Cref{Theo:koopmanbdintegration} provides a generalization of \Cref{Corolla:upperbound_inj_weight}, potentially with a modification to the multiplicative constant. Indeed, for $l = L$, we have\\ $\eR\left(\F^{\,l'+1 \mathord{:} L}\right) = 0$ and 
\begin{align*}
\inf_{{h}' \in \F^{\,l'+1 \mathord{:} L}} &\RE^{\frac{1}{2}}\left[\sup_{h'' \in \F^{\,l'+1 \mathord{:} L}}\left\|h' - \frac{\gamma_n\|h''\|_{H^{s_{l'}}\left(\R^{d_{l'}},\R^m\right)}}{{\|\Tilde{\upsilon}}_n\|_{H^{s_{l'}}\left(\R^{d_{l'}},\R^m\right)}}\Tilde{\upsilon}_n\right\|^2_{H^{s_{l'}}\left(\R^{d_{l'}},\R^m\right)}\right] \\
&\leq \|g\|_{H^{s_L}\left(\R^{d_L},\R^m\right)} 
\RE^{\frac{1}{2}}\left[(1 + \gamma_n)^2\right].  
\end{align*}
\begin{remark}\label{rem:combinedKoopman}
For simplicity, our analysis considers single-output neural networks and RKHSs associated with the one-dimensional Brownian kernel as the function spaces presented in Remark $(2)$-$(i)$ in \cite{Mohammadigohari2025koopman}. A further refinement involves combining the Koopman-based framework with established ``\textit{peeling}'' techniques \cite{neyshabur2015normbased,golowich2018sizeindependent}. This combination has the potential to yield a complexity bound of the form
\begin{align*}
\O\left( \prod_{j=l+1}^L \|\b W_j\|_{2,2} \prod_{j=1}^l \|\b W_j\|\right).    
\end{align*}
This strategy is particularly suitable for many practical network architectures, where layer widths typically exhibit sequential growth near the input ($d_{j-1} \leq d_j$ for small $j$) and decay near the output ($d_{j-1} \geq d_j$ for large $j$), enabling the derivation of tighter bounds relative to existing approaches.    
\end{remark}
\subsection{Excess risk of p-sparsified sketched estimator}\label{subsec:p_sparsified_estimator}
This section presents excess risk bounds for multi-task regression using sketched kernel machines with generic Lipschitz losses.
Let $(\b x_i, \b y_i)_{i=1}^n \in (\mX, \R^m)^n$ represent a dataset drawn from an input space $\mX$ and an output space $\R^m$. Furthermore, assume a decomposable kernel $K_{s_0} = k_{s_0}\b M$ is given, inducing a vvRKHS denoted by $H^{s_0}$. The penalized empirical risk minimization problem is then defined as
\begin{align}\label{sec5:erm}
\min_{f \in H^{s_0}} \frac{1}{n} \sum_{i=1}^n \ell(f(\b x_i), \b y_i) + \frac{\lambda_n}{2} \|f\|_{H^{s_0}}^2,
\end{align}
where $\ell : \R^m \times \R^m \to \R$ represents a loss function satisfying the following conditions: for all $\b y \in \R^m$, the mapping $\b z \mapsto \ell(\b z, \b y)$ is proper, lower semi-continuous, and convex. According to the vector-valued representer theorem \cite{micchelli2005learning}, the minimizer of Problem \eqref{sec5:erm} admits the representation $\hat{f}_n = \sum_{j=1}^n K_{s_0}(\cdot, \b x_j)\hat{\boldsymbol{\alpha}}j = \sum_{j=1}^n k_{s_0}(\cdot, \b x_j)\b M\hat{\boldsymbol{\alpha}}_j$, where the coefficient matrix $\hat{\b A} = (\hat{\boldsymbol{\alpha}}_1, \ldots, \hat{\boldsymbol{\alpha}}_n)^\top \in \R^{n \times m}$ solves the optimization problem
\begin{align*}
\min_{\b A \in \R^{n \times m}} \quad \frac{1}{n} \sum_{i=1}^n \ell([\Gk_0 \b A\b M]^\top_{i:}, \b y_i) + \frac{\lambda_n}{2} \operatorname{Tr}(\Gk_0 \b A\b M\b A^\top).
\end{align*}

In this framework, sketching entails replacing the matrix $\b A$ with a sketched approximation $\b S^\top \b \Gamma$, where $\b S \in \R^{s \times n}$ denotes a sketch matrix and $\b \Gamma \in \R^{s \times m}$ parameterizes the reduced-dimensional representation. Consequently, the solution to the sketched problem is given by $\tilde{f}_s = \sum_{j=1}^n k(\cdot, \b x_j)\b M[\b S^\top \tilde{\b \Gamma}]_{j:}$, with $\tilde{\b \Gamma} \in \R^{s \times m}$ obtained by minimizing

\begin{align*}
\frac{1}{n} \sum_{i=1}^n \ell([\Gk_0 \b S^\top \tilde{\b \Gamma}\b M]_{i:}, \b y_i) + \frac{\lambda_n}{2} \operatorname{Tr}(\b S\Gk_0 \b S^\top \tilde{\b \Gamma}\b M\tilde{\b \Gamma}^\top).
\end{align*}
We adopt the framework of (\cite{li2021towards}, Sections $2.1$ \& $3$) to derive our excess risk bounds. Specifically, we assume that the true risk is minimized over $H^{s_0}$ at $f_{H^{s_0}} \coloneqq \arg \min_{f \in H^{s_0}} \RE[\ell(f(X), Y)]$. The existence of such a minimizer is a standard assumption \cite{caponnetto2007optimal,elahmad2023fast,rudi2017generalization,yang2017randomized} and implies that $f_{H^{s_0}}$ has bounded norm (\cite{elahmad2023fast,rudi2017generalization}, Remark $2$). Following \cite{elahmad2023fast,li2021towards}, we also assume that the estimators obtained by Empirical Risk Minimization possess bounded norms.

Let $\b k_0/n = \b U \b D \b U^\top$ represent the eigendecomposition of the scaled Gram matrix $\b k_0$, where $\b D = \operatorname{diag}(\mu_1, \ldots, \mu_n)$ contains the eigenvalues of $\b k_0/n$ arranged in decreasing order. Define $\delta_n^2$ as the critical radius of $\b k_0/n$, characterized as the minimal value for which the function $\psi(\delta_n) = \left( \frac{1}{n} \sum_{i=1}^n \min(\delta_n^2, \mu_i) \right)^{1/2}$ is bounded above by $\delta_n^2$. The existence and uniqueness of $\delta_n^2$ are well-established for any RKHS associated with a positive definite kernel, as demonstrated in \cite{yang2017randomized}. The statistical dimension $d_n$ is defined in \cite{elahmad2023fast} as the minimal index $j \in \{1, \ldots, n\}$ such that $\mu_j \leq \delta_n^2$, with $d_n = n$ if no such $j$ exists.

\begin{definition}\emph{($\b k_0$-satisfiability, \cite{yang2017randomized})}
Let $ c > 0 $ be independent of $ n $, $ \b U_1 \in \mathbb{R}^{n \times d_n} $ and $\b U_2 \in \mathbb{R}^{n \times (n - d_n)} $ be the left and right blocks of the matrix $ \b U $ previously defined, and $ \b D_2 = \operatorname{diag}(\mu_{d_n + 1}, \ldots, \mu_n) $. A matrix $ \b S $ is said to be \emph{$\b k_0$-satisfiable} for $ c $ if we have
\begin{equation}
\label{eq:k-satisfiability}
\begin{aligned}
\| (\b S \b U_1)^\top \b S \b U_1 - \b I_{d_n} \| &\leq \tfrac{1}{2}, \quad \text{and}\,\,
\| \b S \b U_2 \b D_2^{1/2} \| &\leq c \delta_n.
\end{aligned}
\end{equation}
\end{definition}
A matrix is considered $\b k_0$-satisfiable if it approximately preserves distances among the dominant eigenvectors of $\b k_0$ while exhibiting a limited influence on its minor eigenvectors.

We make the following assumption we will use throughout this section.
\begin{assumption} \label{assum:sketch_satisfiable}
The sketch $\b S$ is $\b k_0$-satisfiable with constant $c > 0$.
\end{assumption}

The following theorem's proof directly applies Theorem $4$ from \cite{elahmad2023fast}, a key theoretical result underpinning this paper.
\begin{corollary}
Suppose that Assumptions \ref{assum:true_risk_min} to \ref{assum:sketch_satisfiable} hold, that $K_{s_0} = k_{s_0}\b M$ is a decomposable kernel with $\b M$ invertible, and let $C = 1 + \sqrt{6}c$,  with c the constant from \Cref{assum:sketch_satisfiable}. Then for any $\delta \in (0, 1)$, with probability at least $1 - \delta$, we have
\begin{align}
\mR\left(\tilde{f}_s\right) \leq \mR\left(f_{H^{s_0}}\right) + J_{\ell}C \sqrt{\lambda_n + \|\b M\| \delta_n^2 }+ \frac{\lambda_n}{2} +  8L \sqrt{\frac{\kappa \Tr(\b M)}{n}} + 2 \sqrt{\frac{8 \log(4/\delta)}{n}}.
\end{align}
\end{corollary}
\cite{elahmad2023fast} proposed $p$-sparsified sketches, demonstrating their $\b k_0$-satisfiability. These sketches consist of i.i.d. Rademacher or centered Gaussian entries modulated by Bernoulli variables (parameter $p$, scaled for isometry). Sparsity, controlled by $p$, enables rewriting the sketch matrix $\b S$ as a product of a sub-Gaussian and a reduced sub-sampling sketch, improving computational efficiency. $p$-sparsified sketches are $\b k_0$-satisfiable with high probability for $c = \frac{2}{\sqrt{p}} \left(1 +  \sqrt{\log\left(5\right)}\right) + 1$ (see, \cite{elahmad2023fast}, Theorem 5).
\begin{figure}[htbp]
  \centering
  \includegraphics[width=0.8\linewidth]{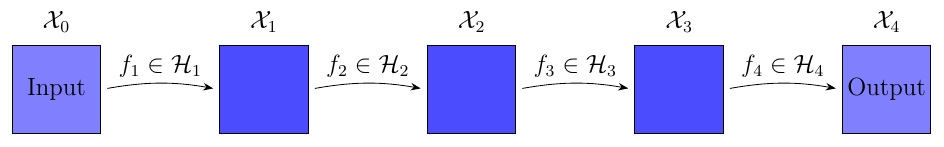}
  \caption{Illustration of the proposed deep vvRKHS (adapted from \cite{hashimoto2023deep}, Figure 1). For the autoencoder, the encoder is represented by $f_1 \circ f_2$, and the decoder by $f_3 \circ f_4$.}
  \label{fig:myfig}
\end{figure}
\section{Operator Representation framework for deep vvRKHS}\label{sec:operator_vvRKHS}
Inspired by recent advances in operator-theoretic approaches to deep learning, particularly the framework developed by \cite{hashimoto2023deep}, this section presents a new network architecture designed to capture intricate relationships in multi-task learning. By strategically employing PF operators within a deep vvRKHS framework, we create a computationally tractable yet theoretically sound approach to deep kernel methods.
\subsection{Deep vvRKHS}
We define an $L$-layer deep vvRKHS. For an integer $L \geq 3$, let $\mX_0, \mX_1, \dots, \mX_L, \mY$ represent a family of Hilbert spaces. For each layer $j = 1, \dots, L$, let $K_j : \mX_{j-1} \times \mX_{j-1} \to \mB(\mX_j)$ be an $\mX_j$-valued positive definite kernel, and $\mH_j$ be the corresponding vvRKHS. Let $\widetilde{\mH}_j$ be the $\mY$-valued vvRKHS associated with the same kernel; $\mH_j \leq \widetilde{\mH}_j$ for $j = 1, \dots, L$. We define function spaces $\F_L = \{f \in \mH_L \mid \|f\|_{\mH_L} \leq B_L\}$ and $\F_j = \{f \in \mH_j \mid \|\mP_f\| \leq B_j\}$ for $j = 1, \dots, L-1$. For $f \in \mH_j$ with $j = 1, \dots, L-1$, $\mP f$ is the PF operator from $\widetilde{\mH}_j$ to $\widetilde{\mH}_{j+1}$. Assuming these operators are well-defined, the class of deep vvRKHS networks is $\F_L = \{f_L \circ \dots \circ f_1 \mid f_j \in \F_j, j = 1, \dots, L\}$. The architecture of this deep vvRKHS is illustrated in \Cref{fig:myfig}.
\subsection{PF generalization bound for deep vvRKHS}\label{subsec:perron_frobenius_bound}
We establish a generalization bound for deep vvRKHSs.  The bound is derived by bounding the Rademacher complexity. Given sample data $(\b x_i, \b y_i)_{i=1}^n \in (\mX_0, \mY)^n$,  a key step in the proof uses the reproducing property and the PF operator to express the function composition as an inner product: $f_L \circ \dots \circ f_1(\b x)^\T \b y = \langle \phi_L(f_{L-1} \circ \dots \circ f_1(\b x))\b y, f_L \rangle_{\widetilde{\mH}_L} = \langle \mP_{f_{L-1} }\dots \mP_{f_1} \phi(\b x)\b y, f_L \rangle_{\widetilde{\mH}_L}$.
\begin{prop}\label{prop:PFRademacherCbd_deepvvRKHS}
 The Rademacher complexity $\eR(\F_L)$ is bounded as  
 \begin{align}\label{sec5: deepvvRKHDbound}
  \eR(\F_L) \leq \frac{1}{n} \sup_{(f_j \in \F_j)_j} \|\mP_{f_{L-1}} \dots \mP_{f_1} |_{\widetilde{\m{V}}(\b x)} \|\, \|f_L\|_{\mH_L} \left[ \sum_{i=1}^n \Tr(K_1(\b x_i, \b x_i)) \right]^{1/2},  
 \end{align}
 where $\widetilde{\m{V}}(\b x)$ is the closed subspace of $\widetilde{\mH}_1$ spanned by $\phi_1(\b x_1)\b y_1, \dots, \phi_1(\b x_n)\b y_n$.
 \begin{proof}
 We have
 \begin{align}
&\eR(\F_L) = \frac{1}{n}\RE \left[\sup_{\substack{ (f_j \in \F_j})_j}  \left|\sum_{i=1}^n \langle\boldsymbol{\sigma}_i, f(\b x_i)\rangle_{\mY}\right|\right] \nonumber\\
&= \frac{1}{n}\RE \left[\sup_{\substack{ (f_j \in \F_j})_j}  \left|\sum_{i=1}^n  \left\langle\mP_{f_{L-1}} \dots \mP_{f_1} \phi_1(\b x_i)\boldsymbol{\sigma}_i,f_L\right\rangle_{\mH_L} \right|\right]\nonumber\\
&= \frac{1}{n}\RE \left[\sup_{\substack{ (f_j \in \F_j})_j}  \left|  \left\langle \mP_{f_{L-1}} \dots \mP_{f_1} \sum_{i=1}^n\phi_1(\b x_i)\boldsymbol{\sigma}_i,f_L\right\rangle_{\widetilde{\mH}_L} \right|\right]\nonumber\\
&\leq \frac{1}{n} \sup_{(f_j \in \F_j)_j} \|\mP_{f_{L-1}} \dots \mP_{f_1} |_{\widetilde{\m{V}}(\b x)} \|\, \|f_L\|_{\mH_L}\RE\left[\left\|\sum_{i=1}^n\phi_1(\b x_i)\boldsymbol{\sigma}_i\right\|_{\mH_1}\right].\label{th3:ineq1}
 \end{align}
where the inequality \eqref{th3:ineq1} is by the Cauchy–Schwartz inequality, By Jensen's inequality, therefore, the inequality of the statement is true. 
\hfill$\square$
\end{proof}
\end{prop}
\begin{remark} \label{rem: deepvvRKHS_with_separable_kernels}
We construct an $L$-layer deep vvRKHS with separable kernels. For $L \geq 3$, let $\mX_0, \mX_1, \dots, \mX_L$ be a collection of Hilbert spaces. For $j = 1, \dots, L$, let $\b M \in \mBO(\mY)^+$ and $\b M_j \in \mBO(\mX_j)^+$ such that $\b M_j \leq \b M$. For each $j = 1, \dots, L$, let $\mH_j$ denote the vvRKHS corresponding to the separable kernel $K_j = k_j \b M_j$, where the $k_j$ are scalar-valued translation-invariant kernels.  Define $\widetilde{\mH}_j$ as the vvRKHS corresponding to the separable kernel $K_j = k_j \b M$. Then, for $j = 1, \dots, L$, we have $\mH_j \leq \widetilde{\mH}_j$, as shown in (\cite{JMLR:v13:zhang12a}, Proposition 17). Consequently, the Rademacher complexity can be bounded as
  \begin{align}\label{rem: deepvvRKHS_bound_SepKer}
\eR(\F_L) \leq \frac{\sqrt{\kappa\Tr(\b M_1)}}{n} \sup_{(f_j \in \mathcal{F}_j)_j} \|\mP_{f_{L-1}} \dots \mP_{f_1} |_{\widetilde{\m{V}}(\b x)} \|  \|f_L\|_{\mH_L},
 \end{align}
where we assume $k_1$ satisfies \Cref{assum:kernel_bound}.
\end{remark}
\begin{remark}\label{rem: deepvvRKHS_with_sep_ker_refine}
In \Cref{rem: deepvvRKHS_with_separable_kernels}, let $\b M_1 = \dots = \b M_L =\b M\in \bH^+$ and for $f \in \mH_j$ with $j = 1, \dots, L-1$, $\mP_f: \mH_j \to \mH_{j+1}$ is the PF operator.  In kernel methods for multi-task learning, the selection of the vector-valued reproducing kernel $K_L$ significantly impacts model performance. Given sample data $z = (\b x_i, \b y_i)_{i=1}^n \subseteq \mX \times \R^m$, the minimizer $f_z$ of the following minimization problem, 
\begin{equation}\label{rem: deepvvRKHS_with_sep_ker_refine_minimizer}
    \min_{(f_j \in \mathcal{F}_j)_j} \frac{1}{n} \sum_{i=1}^{n} \|f(\b x_i) - \b y_i\|_2^2 + \lambda_1\|\mP_{f_{L-1}} \dots \mP_{f_1} \| + \lambda_2\|f_L\|_{\mH_L},
\end{equation}
provides a function $f$ within the vvRKHS $\mH_L$. However, a kernel with large capacity (overfitting) forces $f_z$ for the minimizer of \eqref{rem: deepvvRKHS_with_sep_ker_refine_minimizer} to over-adapt to noise. Refinement addresses this by seeking a kernel $G^{(L)}$ corresponding to the vvRKHS $\mH^{(L)}$ such that $\mH^{(L)} \subset \mH_L$ (as function spaces, we consider vvRKHSs $\mH^{(j)}$ corresponding to the separable kernels $G^{(j)}  = k_j\b A$, where $\b A \leq \b M$ for $\b A \in \bH^+$ and $j = 1, \dots, L$), effectively constraining the solution space and reducing the sampling error. Conversely, if $\mH_L$ is too small (underfitting), the minimizer obtained
from \eqref{rem: deepvvRKHS_with_sep_ker_refine_minimizer} cannot accurately capture the underlying function dependency. In this case, a refinement $G^{(L)}$ with $\mH_L \subset \mH^{(L)}$ (as function spaces, we consider vvRKHSs $\mH^{(j)}$ corresponding to the separable kernels $G^{(j)}  = k_j\b A$, where $\b M \leq \b A$ for $\b A \in \bH^+$ and $j = 1, \dots, L$) enlarges the candidate function space, diminishing approximation error. By appropriately refining the vvRKHS, we modulate the trade-off between approximation and sampling errors, mitigating both overfitting and underfitting \cite{JMLR:v13:zhang12a}.
\end{remark}

\section{Conclusion}\label{sec6}
This paper introduces novel generalization bounds for deep learning through operator-theoretic frameworks. For neural networks, we leverage Koopman operators, providing a new perspective on the generalization of high-rank weights and enabling layer-wise analysis. While our bound suggests a potential avenue for regularization, future research should explore connections between our analysis and existing studies of learning dynamics with Koopman operators. We also propose a deep vvRKHS framework analyzed through PF operators. This yields a generalization bound with connections to benign overfitting. 
Extending this work, future research will encompass several key directions. First, empirical validation of our framework with Lipschitz-continuous losses in multi-task deep robust and quantile regression settings will further demonstrate the practical impact of these findings. Second, we aim to generalize our analysis beyond separable kernels and address the well-definedness of Perron-Frobenius operators for broader kernel classes, enhancing the theoretical scope and applicability of our approach.
%
%
%
%

\bibliographystyle{splncs04}
\bibliography{sn-bib}

\end{document}